\documentclass{article}
\usepackage[preprint]{log_2023}

\usepackage{algorithm}
\usepackage{amssymb}
\usepackage{algpseudocode}
\usepackage{caption}
\usepackage{xspace}

\usepackage{xcolor}
\usepackage{amsmath}
\usepackage{amsfonts} 
\usepackage{float}
\usepackage[backref=page]{hyperref}

\newcommand{\real}{{\rm I\!R}}
\newcommand{\bw}{\mathbf{w}}
\newcommand{\method}{\textsc{MP-GNN}\xspace}

\usepackage[numbers,compress,sort]{natbib}

\usepackage{verbatim}
\usepackage{subcaption}

\usepackage{graphicx}
\usepackage{booktabs}            
\usepackage{multirow}            
\usepackage{amsfonts}            
\usepackage{graphicx}            
\usepackage{duckuments}          
\usepackage{adjustbox}

\usepackage[numbers,compress,sort]{natbib}

\title[Meta-Path Learning for Multi-relational Graph Neural Networks]{Meta-Path Learning for Multi-relational Graph Neural Networks}

\author[Francesco Ferrini et al.]{%
Francesco Ferrini\\
\institute{University of Trento, Italy}\\
\email{francesco.ferrini@unitn.it}\And
Antonio Longa\\
\institute{University of Trento, Italy}\\
\email{antonio.longa@unitn.it}\And
Andrea Passerini\\
\institute{University of Trento, Italy}\\
\email{andrea.passerini@unitn.it}\And
Manfred Jaeger\\
\institute{Aalborg University, Denmark}\\
\email{jaeger@cs.aau.dk}
}

\begin{document}
\maketitle

\begin{abstract}
Existing multi-relational graph neural networks use one of two strategies for identifying informative relations:  either they reduce this problem to low-level weight learning, or they rely on handcrafted chains of relational dependencies, called meta-paths. However, the former approach faces challenges in the presence of many relations (e.g., knowledge graphs), while the latter requires substantial domain expertise to identify relevant meta-paths.
In this work we propose a novel approach to learn meta-paths and meta-path GNNs that
are highly accurate based on a small number of informative meta-paths. 
Key element of our approach is a scoring function for 
measuring the potential informativeness
of a  relation in the incremental construction of the meta-path.  
Our experimental evaluation shows that the approach manages to correctly identify relevant meta-paths even with a large number of relations, and substantially outperforms existing multi-relational GNNs on synthetic and real-world experiments.
\end{abstract}

\section{Introduction}
Graph neural networks (GNNs) have emerged as a powerful framework for analyzing networked data \cite{scarselli2008graph, 1555942, henaff2015deep, 4773279}, enabling effective learning and representation of complex relationships in several real-world applications~\cite{zhang2021graph,buffelli2021attention,samoaa2022tep,vinas2022hypergraph}. 
Standard GNN approaches have mostly focused on homogeneous 
graphs~\cite{xu2018powerful,velivckovic2017graph,hamilton2017inductive}, where all nodes and edges belong to a single type. However, many real-world graph datasets exhibit heterogeneity, with multiple types of nodes and relations \cite{sun2017crosslingual, 10.1145/3097983.3098036,ringler2017one}.

Treating heterogeneous graphs as homogeneous and aggregating information uniformly across all relations is a suboptimal approach, as different relations can convey largely different semantic information about the nodes they connect. 
A simple and effective strategy to retain the rich semantic information present in heterogeneous graphs is relying on meta-paths, which are chains of relational dependencies (e.g., "actor -> acted in -> movie -> has genre").
The challenge lies in determining the relevant meta-paths in a given graph. Existing methods either rely on predefined meta-paths defined by domain experts \cite{Fu_2020, CHANG2022107611, wang2021heterogeneous}, which are extremely expensive to collect, or learn "soft" meta-paths by learning to assign weights to relations \cite{schlichtkrull2017modeling, yun2020graph, yun2021graph}, an approach that only works with few relations and fails to scale to knowledge graphs.
Solutions conceived for mining meta-paths from knowledge graphs typically consider relations only, ignoring node features altogether~\cite{10.5555/3367471.3367477,xiong2018deeppath}.

To overcome these limitations, we propose a novel approach to learn meta-paths and meta-path GNNs that
are highly accurate based on a small number of informative meta-paths. Key to our approach is the formalization of a scoring function, inspired by the relational information gain principle \cite{RIG}, that evaluates the potential informativeness of a relation in the incremental construction of the meta-path. This allows to learn a Meta-Path Graph Neural Network (MP-GNN) in which 
different layers convey information from different relations while retaining node-specific features in the aggregation process.

The main contributions of this work can be summarized as follows:
\begin{itemize}
    \item We propose a scoring function evaluating the potential informativeness of a relation in the meta-path construction.
    \item We introduce MP-GNN, a simple variant of the RGCN architecture, which effectively combines learned meta-paths and node features into a multi-relational graph processing architecture.
    \item We provide an extensive experimental evaluation on synthetic and real-world datasets showing how our approach substantially outperforms existing multi-relational GNNs when dealing with graphs with a large number of relations.
    \end{itemize}

\section{Related work}

In recent research, meta-path mining has emerged as an effective approach for analyzing heterogeneous graphs, relying on frequency cutoffs and sequential pattern mining strategies to identify promising meta-paths~\cite{7022636, Discovering_Meta-Paths_in_Large_Heterogeneous_Information_Networks, Yu2012UserGE}. In the field of neuro-symbolic reasoning for knowledge graph completion (KGC), various approaches use reinforcement learning-based algorithms to explore relation-paths and derive logical formulas \cite{10.5555/3367471.3367477, xiong2018deeppath}. Other approaches \cite{taocohen, lao-etal-2012-reading, neelakantan-etal-2015-compositional}, search for the most relevant meta-path using variants of random walk search. A major limitation of all these approaches is that they are incapable of accounting for node features in determining the relevance of a candidate meta-path, making them unusable in knowledge graph embedding scenarios.

In the field of heterogeneous graph embedding, several methods have been proposed to enhance node and graph embedding by incorporating meta-paths. These methods can be broadly categorized into two groups: those using predefined meta-paths and those learning meta-paths by weighting the contribution of different relations.

In the first group, Meta-path Aggregated Graph Neural Network \cite{Fu_2020} focuses on aggregating node features along predefined meta-paths using an attention mechanism, capturing diverse structural patterns. Heterogeneous Attention Network \cite{wang2021heterogeneous} introduces a hierarchical attention mechanism to handle heterogeneous graphs, enhancing performance and interpretability. GraphMSE \cite{Li_Jin_Song_Zhu_Shi_Wang_2021} tackles the problem of information-rich meta-path selection by aggregating information using different meta-paths and adopting BFS (Breadth First Search) as meta-path expansion strategy. Meta-path extracted graph neural network  \cite{CHANG2022107611} incorporates message passing and emphasizes interpretability and semantic relationships. 
However, these approaches require that meta-paths are provided beforehand, something which severely limits their adaptability.

In the second group, Relational Graph Convolutional Networks (RGCN) \cite{schlichtkrull2017modeling} capture relation-specific patterns with distinct trainable parameters for each edge type. R-HGNN \cite{RHGNN} uses a dedicated graph convolution component for unique node representations from relation-specific graphs. RSHN \cite{RSHN} integrates Coarsened Line Graph Neural Network (CL-GNN) for enhanced embedding in large-scale heterogeneous graphs. Graph Transformer Networks (GTN) \cite{yun2020graph} learn task-specific multi-hop connections (i.e., meta-paths) for useful node representations. FastGTN \cite{yun2021graph} addresses GTN's high complexity by implicitly transforming graphs. HGN \cite{HGN} employs GAT as a backbone for a simple yet effective model. HGT \cite{HGT} uses node and edge-type dependent parameters for heterogeneous attention. MVHRE \cite{MVHRE} enriches relational representations for link prediction using a multi-view network representation learning framework. While effective with a small number of candidate relations, these approaches' performance degrades as the number increases, as shown in our experimental evaluation.

\section{Preliminary}
In this section, we provide an overview of fundamental concepts 
of our approach.

\textbf{Heterogeneous graph.} 
A heterogeneous graph is a directed graph ${\cal G} = ({\cal V}, {\cal E}, {\cal T}_v, {\cal T}_e)$ where ${\cal V}$ is the set of nodes or entities and ${\cal E}$ is the set of edges. Each node $v$ and edge $e$ has a type, specified by the mapping functions
$\tau_v(v): {\cal V} \xrightarrow{}  {\cal T}_v$ and $\tau_e(e): {\cal E} \xrightarrow{}  {\cal T}_e$.
Moreover, each node $v$ has a feature vector $x_v \in \real^d$.

\textbf{Meta-path} A meta-path $mp$ is a relation sequence defined on a heterogeneous graph $\cal G$, denoted in the form $\xrightarrow{r_1} \xrightarrow{r_2} ... \xrightarrow{r_L}$, where $r_1, ..., r_L$ are relation types and for each consecutive pair of relations $\xrightarrow{r_i} \xrightarrow{r_{i+1}}$ the intersection between the valid node types that are the destination of $\xrightarrow{r_i}$ and the valid node types that are the source of $\xrightarrow{r_{i+1}}$ is non-empty. Note that this is a more general definition than the one in~\cite{metapath}, in that it allows multiple node types as sources and destinations of a given relation, consistently with what can be found in large general-purpose knowledge graphs. 

\textbf{RGCN layer} The relational graph convolutional layer from \cite{schlichtkrull2017modeling} extends the standard convolution operation on graphs \cite{kipf2017semisupervised} to the multi-relational setting by assigning specific parameters for each relation type.
Message passing update for node $i$ at layer $l$ is given by:

\begin{equation}
  h^{(l+1)}_i = \sigma\left(W_0^{(l)} h^{(l)}_i +  \sum_{r \in {\cal R}} \sum_{j\in {\cal N}_i^{r}} \frac{1}{c_{i,r}} W_r^{(l)}h^{(l)}_j   \right)
  \label{formula:rgcn}
\end{equation}
where $\cal R$ is the set of relations in the graph, ${\cal N}_i^{r}$ is the set of $r$-neighbours of node $i$ and $c_{i,r}$ is a fixed or learnable normalizing parameter.

\section{MP-GNN learning}

\begin{figure}[t!]
    \centering
    \includegraphics[width = 0.9\textwidth]{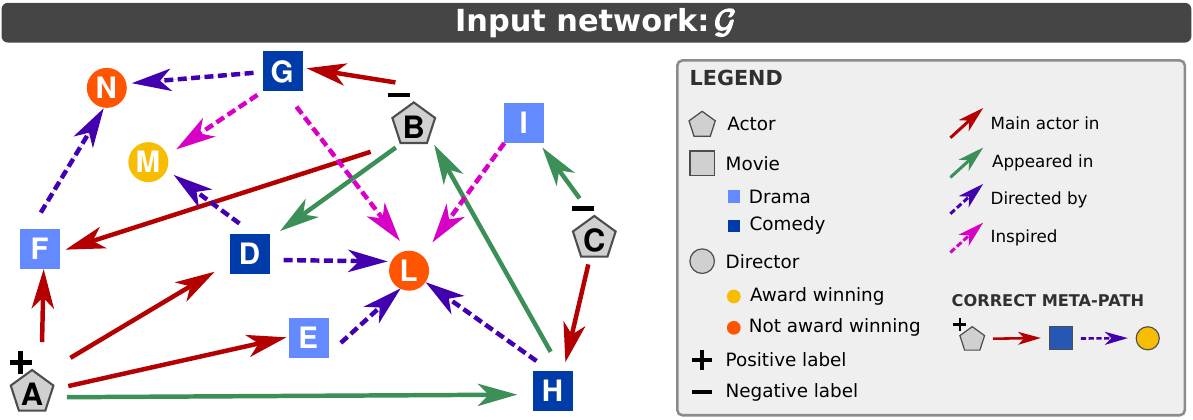}
    \caption{A toy node classification problem. Node shapes indicate types, while node attributes and edge types are encoded as colors. The task consist in labelling actor nodes (pentagons, which do not have attributes). An $Actor$ is labelled as positive if involved as main actor in a drama directed by and award winning director.}
    \label{fig:inputnetwork}
\end{figure}

The goal of our approach is to learn relevant meta-paths that can serve as predictive
features for the node classification task.~\footnote{We focused on node classification in this paper, but the approach can be adapted to deal with graph classification tasks, e.g., by considering a supernode connected to all graph nodes as the target.} Differently from the approaches
that use all relations at the same time by weighting each edge type contribution,
we focus on finding the relevant chains of relations (i.e., meta-paths) beneficial for making accurate predictions. Note that, differently from what happens in purely relational settings \cite{7022636, Discovering_Meta-Paths_in_Large_Heterogeneous_Information_Networks, Yu2012UserGE, taocohen, lao-etal-2012-reading, neelakantan-etal-2015-compositional}, we assume here that the informativeness of a meta-path also depends on the features of the nodes that are being traversed (which include the node type, but also node attributes and potentially pre-computed node embeddings). Our approach accounts for this aspect in mining relevant meta-paths. Meta-paths are constructed in a greedy, incremental approach using the  idea of relational information gain \cite{RIG} to score candidate extensions of an already constructed meta-path prefix. 
Consider the toy node classification task in Figure~\ref{fig:inputnetwork}. To incrementally build the correct meta-path (bottom right in the legend), one has to realize that "Main actor in" is a better candidate than "Appeared in". Intuitively, our scoring function does this by assigning weights (i.e., pseudo-labels) to nodes reached by a candidate relation in such a way that the label of the target node can be inferred by propagating the pseudo-label of the neighbour. Figure~\ref{fig:firstiterationscore} shows an example of weight assignment for the "Main actor in" and "Appeared in" relations, indicating a higher score for the former. However, these pseudo-labels only hint at the potential informativeness of the relation. Indeed, being a main actor in a movie is not enough to qualify as an award winning actor, even in the toy example of Figure~\ref{fig:inputnetwork}. The movie should be a drama (node feature), and be directed by an award winning director. Whether this potential informativeness actually materializes is determined in the following steps, where the pseudo-labels become new prediction targets for the next extension of the meta-path under construction. Details of this method are described in Section \ref{scorefunctiondescription}.

Once a candidate meta-path has been extracted, it is used to build a MP-GNN in which each layer corresponds to a relation on the meta-path. Section~\ref{MP-GNN} presents a formalization of this architecture, and shows how to extend it to account for multiple meta-paths. Finally, these ingredients are combined into an overall algorithm to jointly learn a meta-path and a corresponding MP-GNN. For the sake of readability, the algorithm is presented for the single meta-path case, but its extension to multiple meta-paths using a beam search is straightforward (we employed a beam size equal to three in the experiments). Note that this algorithm is designed to identify existential meta-path features, i.e., cases where the existence of an instance of a meta-path is informative for the class label. Adaptations and extensions where counts or proportions of meta-path realizations are the relevant feature are subject of future work.

\subsection{Scoring function}
\label{scorefunctiondescription}

The goal of the scoring function is that of providing a measure of informativeness of a relation towards predicting node labels. We start discussing the first iteration, i.e., identifying the first relation in the meta-path, and then show how the function can be adapted to deal with meta-path extension.

\begin{figure}[ht]
	\centering
        \includegraphics[width=0.9\linewidth]{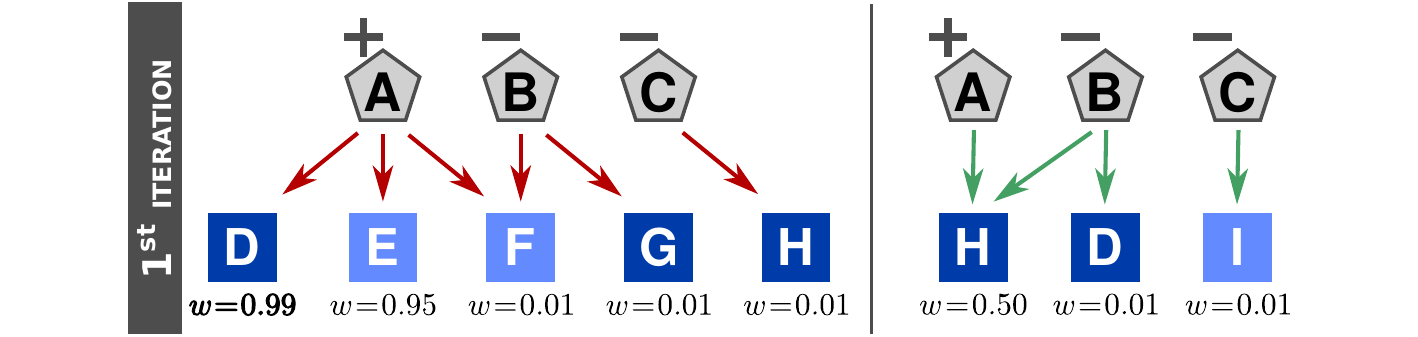}
	\caption{\textbf{First iteration:} the scoring function assigns a high score to the red ("Main actor in") relation (left panel) by giving large weights to movies D and E, that are only connected to a positive node, and small weights to the other movie nodes. On the other hand, the green ("Appeared in") relation (right panel) has low score, as no weight assignment can jointly explain the positive label of the A node and the negative label of the B node.
 }
	\label{fig:firstiterationscore}
\end{figure}

In the first iteration, the scoring function takes as input a list of nodes together with their target labels. Under the previously introduced existential quantification assumption, a candidate relation $r$ is informative for the label of a node $i$ if at least one of the neighbors ${\cal N}_i^r$ of $i$ according to $r$ belongs to the ground-truth meta-path, and $i$ has the right features (remember that the label is assumed to depend on the combination of the meta-path and the features of the nodes being traversed). This can be formalized as follows:
\begin{equation}
\label{eq:sf_first_iter}
    \tilde{y}_i^r = \Theta^T h^{(0)}_i \cdot \max_{j \in {\cal N}_i^r} w_j
\end{equation}
Here $\Theta$ is a learnable weight vector accounting for the contribution of the node features, while $w_j$ is a learnable node weight that is set (close) to 1 if node $j$ is predicted as belonging to the ground-truth meta-path, and (close to) zero otherwise. The score of $r$ is computed by minimizing the MSE between the predicted and ground truth node labels over $\Theta$ and $\bw$:
\begin{equation}
\label{eq:sf_loss}
    s_r = \min_{\Theta,\bw} \dfrac{1}{N} \sum_{i=1}^N (\tilde{y}_i^r - y_i)^2
\end{equation}
The relation with the minimum score is selected as the first relation of the meta-path. 

\begin{figure}[ht]
	\centering
        \includegraphics[width=0.9\linewidth]{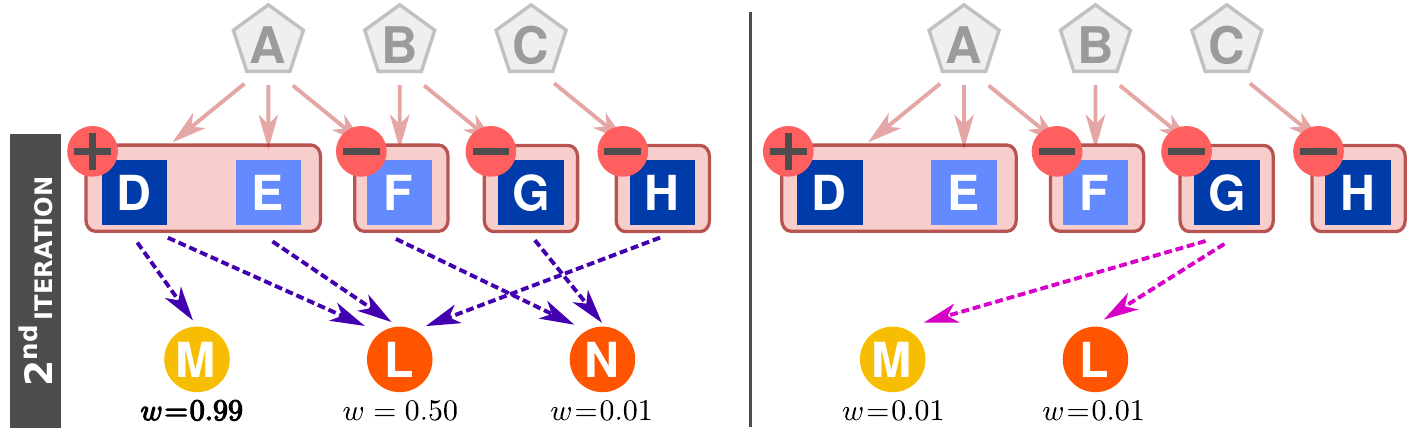}
	\caption{\textbf{Second iteration}: The scoring function assigns a high score to the purple ("Directed by") relation (left panel) by assigning a large weight to the M director, which is only one connected to a positive bag, and small weights to the other directors. On the other hand, the "pink" (Inspired) relation (right panel) gets a low score as no weight assignment is compatible with the positive bag.
 }
	\label{fig:second_iteration_score}
\end{figure}

To explain how the scoring of the following relations in the meta-path works, it is important to remember that the weights $\bw$ represent a tentative assignment to neighbours as belonging or not-belonging to the ground-truth meta-path (i.e., their {\em potential informativeness}). Multiple potential assignments can be minimizers of Eq.~\ref{eq:sf_loss}. In the left panel of Figure~\ref{fig:firstiterationscore}, where relation $r_1$ (green) is being scored, any minimizers of Eq.~\ref{eq:sf_loss} requires $w_E=1$ (to account for the positive label of node $A$) and $W_F=0$ (to account for the negative label of node $B$). On the other hand, (0,1), (1,0) and (1,1) are all valid assignments to the $(W_G,W_A)$ pair. Indeed, the only constraint that the positive label of $C$ enforces is that the bag $(W_G,W_A)$ contains at least one node with value 1, as happens in multi-instance classification settings~\cite{AMORES201381}. We thus generate labelled bags of nodes for the following iteration(s) of meta-path construction, that will play the role of the node labels $y$ in the initial iteration. Positive bags are computed as follows:
\begin{equation}
\label{eq:pos_bags}
B^+(i) = \{ j \in {\cal N}_i^r  \,|\, \nexists k : j \in {\cal N}_k^r \land y_k = -1\}
\end{equation}
where $i$ is a positive-labelled node ($y_i = +1$). Negative bags, on the other hand, are singletons, i.e., given a negatively-labelled node $j$, we create a negative bag $B^-(k) = \{k\}$ for each of its neighbors $k \in {\cal N}_j^r$. The informativeness of the new relation $s$ (as extension of relation $r$) can now be computed in terms of its potential in predicting bag labels:
\begin{equation}
\label{eq:sf_bags}
\tilde{y}_{B(i)}^{s} = \max_{j \in B(i)} \Bigl( \Theta^T  h^{(0)}_j  \cdot \max_{k\in {\cal N}^s_j} w_k \Bigl)
\end{equation}
and obtained minimizing MSE at the bag-label level. See Figure~\ref{fig:second_iteration_score} for a graphical representation of the components involved.

Once the next relation is selected, the procedure could in principle continue, by further expanding positive bags with a procedure analogous to Eq.~\ref{eq:pos_bags}, where $i$ is itself replaced with a bag of nodes. However, this procedures ends up diluting information too much, so that the informativeness of relations becomes difficult to predict. We rather assign a positive label to a node within a bag if it is used to predict the positive label of the bag (Eq.~\ref{eq:sf_first_iter}) at least once out of $M$ restarts with randomly initialized weights. See the Appendix for the details.

\subsection{MP-GNN}
\label{MP-GNN}

In the single meta-path MP-GNN, a meta-path $mp = r_1, . . . , r_L$ induces a multi-relational GNN with $L$ layers, that we denote by \textsc{MP-GNN}($mp$). The first layer is associated to the last relation of the meta-path $r_L$, and so on until the final layer which is associated with $r_1$. The message passing update is formalized as follows: 
\begin{equation}
\label{formula:MP-GNN}
  h^{(l+1)}_{i} = \sigma\left( W_0^{(l)} h^{(l)}_{i} +  \sum_{j\in {\cal N}_i^{r_{L-l+1}}} \frac{1}{|{\cal N}_i^{r_{L-l+1}}|} W^{(l)}h^{(l)}_{j}   \right)
\end{equation}
where $l$ ranges from 1 to $L$.

The definition can be generalized to deal with multiple meta-paths by concatenating  embeddings coming from the different meta-paths:

\begin{equation}
\label{formula:MP-GNNmultiple}
h^{(l)}_{i} = \mathbin\Big\Vert_{k=1}^K h^{(l)}_{(i,k)}
\end{equation}
where $K$ is the number of meta-paths, $h^{(l)}_{(i,k)}$ is the embedding of node $i$ according to meta-path $k$ computed using Eq.~\ref{formula:MP-GNN} and $\mathbin\Vert$ is the concatenation operator. 

It is worth mentioning here that while this definition of MP-GNN is a straightforward adaptation of the RGCN architecture to deal with learned meta-paths, more complex architectures involving pre-defined meta-paths could in principle be employed~\cite{Fu_2020,wang2021heterogeneous,Li_Jin_Song_Zhu_Shi_Wang_2021,CHANG2022107611}. We opted for this simple choice in the paper so as to best single out the contribution of the scoring function in determining the performance of the resulting architecture.

\subsection{Overall algorithm}

The overall algorithm for learning MP-GNN is outlined in Algorithm \ref{algorithmprincipal}. The algorithm takes as inputs a heterogeneous graph ${\cal G}$, a set of candidate relations ${\cal R}$, a set of node-label pairs $labels$ and a hyper-parameter $L_{MAX}$ indicating the maximal length of candidate meta-paths. The algorithm repeatedly call the scoring function (Eq. \ref{eq:sf_loss}) to score candidate relations and keeps track of the best scoring one. It then builds an MP-GNN with the current (partial) meta-path and trains it to predict node labels, using $F_1$ score (computed on a validation set, omitted in the algorithm for the sake of compactness) as final meta-path evaluation metric. Note that this is the only "real" measure of meta-path quality, as the one computed by the scoring function is still a "potential" informativeness, that only fully materializes when the meta-path is embedded into an MP-GNN. The algorithm keeps track of the highest $F_1$ meta-path prefix so far, and proceeds by generating labelled bags as described in Section~\ref{scorefunctiondescription} for the next round of relation scoring.

As previously mentioned, the algorithm is presented for the sake of simplicity in the single meta-path case. However, the actual implementation performs beam search on the space of meta-paths, retaining the $K$ top-scoring ones according to Eq. \ref{eq:sf_loss} and concatenating their embeddings into the MP-GNN as per Eq.~\ref{formula:MP-GNNmultiple}. Notice that in evaluating the resulting MP-GNN, meta-paths not contributing to increasing $F_1$ are discarded, so as to retain only the informative meta-paths in the final architecture.

\begin{algorithm}
\caption{\textsc{LearnMP-GNN} algorithm. ${\cal G}$ is a heterogeneous graph, ${\cal R}$ is the set of possible relations, $labels$ is the initial set of node-label pairs and $L_{MAX}$ is the maximal meta-path length}
\label{algorithmprincipal}
\begin{algorithmic}[1]
\Procedure{\textsc{LearnMP-GNN}}{${\cal G}$, ${\cal R}, labels, L_{MAX}$}
    \State Initialize $mp^* \leftarrow [\,]$, $mp \leftarrow [\,]$, $F_1^* \leftarrow 0$, $target \gets labels$
    \While{$|mp| < L_{MAX}$}
        \For{$r \in {\cal R}$}
           \State $s_r \leftarrow \textsc{score-relation}({\cal G}, target, r)$ \Comment{Equation \ref{eq:sf_loss}}
        \EndFor
        \State $r^* \gets \textrm{best scoring relation}$   
        \State $mp \leftarrow mp,r^*$       
        \State $gnn \leftarrow \textsc{train}(\textsc{MP-GNN}(mp), {\cal G}, labels)$ 
        \State $F_1 \leftarrow \textsc{test}(gnn)$
        \If{$F_1 > F_1^*$}
            \State $mp^* \leftarrow mp, \; F_1^* \leftarrow F_1$            
        \EndIf
        \State $target \leftarrow \textsc{generate-bags}(target,r^*)$\Comment{Section \ref{scorefunctiondescription}}
    \EndWhile
    \State \textbf{return} $mp^*$ 
\EndProcedure
\end{algorithmic}
\end{algorithm}

\section{Experimental results}
Our experimental evaluation aims to answer the following research questions:

\begin{itemize}
    \item[{\bf Q1}] Can \method recover the correct meta-path for an increasing number of candidate relations? 
    \item[{\bf Q2}] Is \method competitive with existing approaches in real-world datasets with few relations?
    \item[{\bf Q3}] Does \method outperform existing approaches in real-world datasets with many relations?
\end{itemize}

We compared \method with existing solutions that: 1) do not require to pre-specify relevant meta-paths, 2) can handle (possibly high-dimensional) node features. Given these requirements, we identified the following competitors:

\begin{itemize}
    \item {\bf RGCN} \cite{schlichtkrull2017modeling}, a generalization of the GCN architecture to the multi-relational case, that employs a different matrix of parameters for each edge type. 
    \item {\bf GTN} \cite{yun2020graph} can convert an input graph into different  meta-path graphs for specific tasks and learn node representations within these graphs.
    \item {\bf FastGTN} \cite{yun2021graph}, an efficient variant of GTN that avoids adjacency matrices multiplication for graph transformation. 
    
    \item {\bf R-HGNN} \cite{RHGNN}, employs a different convolution for each edge type. Finally combines different embeddings with a cross-relation message passing.
    
    \item  {\bf HGN} \cite{HGN}, utilizes GAT as backbone to design an extremely simple HGNN model.

\end{itemize}

We implemented \method using Pytorch Geometric \cite{paszke2019pytorch}, while the code of the competitors was taken from their respective papers.
For \method we used Adam optimizer with a learning rate of 0.01.  
We set the maximum meta-path length $L_{MAX}=4$ and the beam size $K=3$. 
We used an 80/20/10 split between train, validation and test in all cases, with model selection performed on the validation set for all methods. We employed F1-macro score on the test set as evaluation metric to account for the unbalancing present in many of the datasets.
The code is available at \href{https://github.com/francescoferrini/MultirelationalGNN}{LINK}.

In the following we report the experimental setting and the results we obtained in addressing each of the research questions under investigation. The statistics of the datasets used in the experiments are reported in the Appendix.

\subsection{Q1: \method consistently identifies the correct meta-path}
\label{syntheticresults}

In order to answer the first research question, we designed a controlled setting where the correct meta-path is known, and experiments can be run for an increasing number of candidate relations. We generated synthetic datasets where nodes are typed A or B, the number of relations $|\cal R|$ varies 
in $\{ 4, 8, 10, 14 \}$, and the number of relations that can connect more than one pair of node types
(e.g., $A \stackrel{r_1}{\rightarrow} B$ and $A \stackrel{r_1}{\rightarrow} A$). The ground truth meta-path consists of a (valid) sequence of relations and nodes of a given type (e.g.,  $x \stackrel{r_1}{\rightarrow} A \stackrel{r_3}{\rightarrow} B$, with $x$ being a node of arbitrary type). Nodes are labelled as positive if found to be starting points of a ground-truth meta-path, and negative otherwise. We generated labelled datasets using ground-truth meta-paths of different lenghts $L \in \{2,3,4\}$. 
Details of the different settings can be found in the Appendix (Figure \ref{fig:experimental_setting}). 

\begin{figure}[ht]
	\centering
	\includegraphics[width=\linewidth]{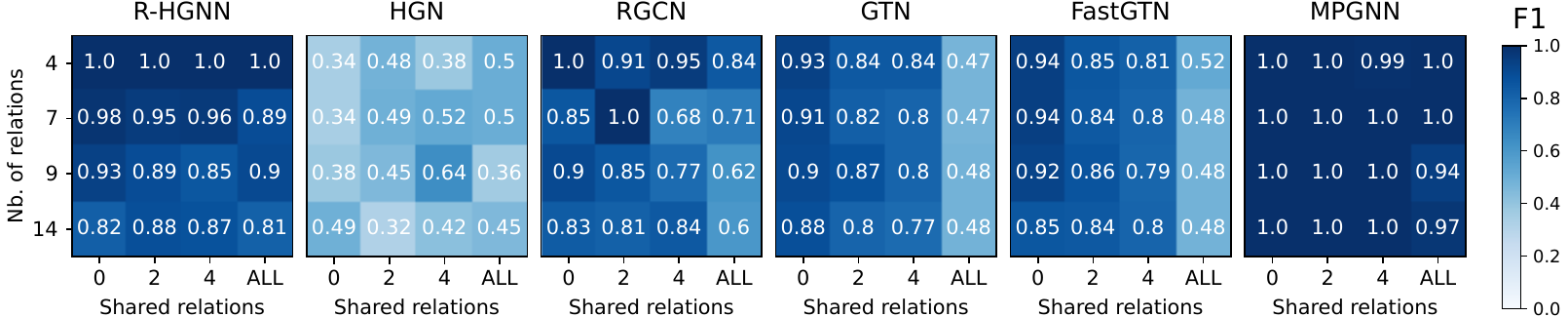}
	\caption{Synthetic setting: F1-score (the darker the better) as a function of the overall number of relations (rows) and the number of shared relations (columns).} 
	\label{fig:matricesres}
\end{figure}

Figure \ref{fig:matricesres} shows the $F_1$ score for each model when varying the overall number of relations and the number of shared relations, for a ground-truth meta-path of length three. Darker cells correspond to higher $F_1$ value. Results show that the performance of existing multi-relational GNN approaches is severely affected by the relational complexity of the graph, with RGCN and R-HGNN being more sensible to the overall number of candidate relations and GTN and FastGTN having bigger problems with the number of shared relations, while HGN has poor performance in all settings, likely due to its lack of an explicit modelling of relation types. Conversely, \method consistently achieves optimal or quasi-optimal performance in all settings. Whenever $F_1=1$, \method manages to perfectly recover the ground-truth meta-path, while values smaller than one are due to spurious relations being added at the end of the meta-path (which however have a limited impact on predictive performance). 

\begin{figure}[ht]
	\centering
	\includegraphics[width=0.9\linewidth]{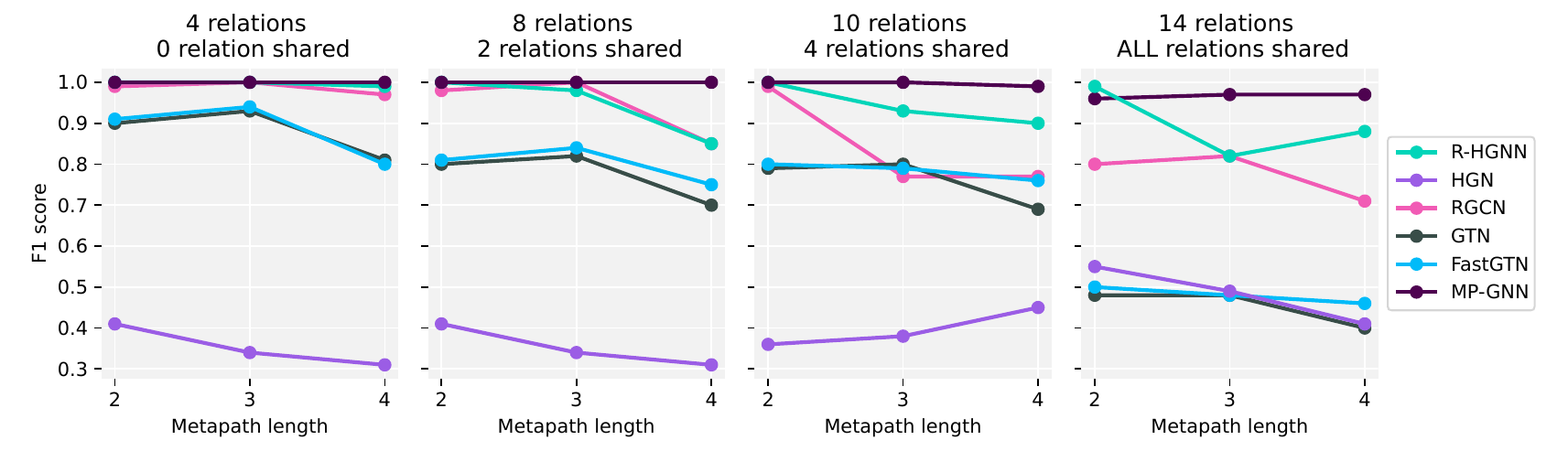}
	\caption{Synthetic setting: F1-score as a function of the ground-truth meta-path length, for an increasing complexity of the search space.}
	\label{fig:metalengthres}
\end{figure}

Figure \ref{fig:metalengthres} shows results when increasing the relational complexity of the network {\em and} the length of the meta-path characterizing the positive class. Each setting corresponds to an entry in the main diagonal of Figure \ref{fig:matricesres}, where we additionally varied the length of the meta-path from 2 to 4. 
Results show that GTN, FastGTN and HGN struggle in most settings, while RGCN and R-HGNN are competitive in the simplest settings (few relations and/or short meta-paths) but its performance quickly degrade when the size of the search space increases. Again, \method consistently achieves excellent performance in all settings, almost always perfectly recovering the ground-truth meta-path.

\subsection{Q2: \method achieves state-of-the-art results on real-world datasets with few relations}
\label{acmresults}

The second set of experiments focuses on popular real-world benchmarks for multi-relational GNNs. In all cases the task is multi-class classification at the node level. We quickly summarize the characteristics of the benchmarks in the following: \textbf{IMDB}: a dataset extracted from the popular Internet Movie Database. It contains 3 types of nodes (movies (M), directors (D) and actors (A)) and uses the genres of movies as labels. \textbf{DBLP}: citation network where nodes are of paper (P), author (A) or conference (C) type,  connected by edge types PA, AP, PC, CP, and the task is predicting the research area of authors. \textbf{ACM}: again a citation network, similar to the one of DBLP with  conference nodes replaced by subject (S) nodes (and edge types replaced accordingly).

\begin{table}[ht]
\begin{minipage}{.30\textwidth}
        \caption{Few-relations datasets. \textbf{(Top)}:  $F_1$ scores, mean and std computed over five runs. Best results highlighted in bold. \textbf{(Bottom)}: learnt meta-paths for \method and GTN/FastGTN (which learn the very same meta-paths). Other baselines not reported as they do not explictly extract meta-paths.}
        \label{tab:acmresults}
 \end{minipage}\hfill\begin{minipage}{.68\textwidth}
    \centering
    \begin{tabular}{llll}
         \hline
        Model   & DBLP & IMDB & ACM  \\ \hline
        R-HGNN & 0.86${\scriptstyle (\pm0.04)}$ & \textbf{0.64}${\scriptstyle (\pm0.01)}$ & 0.9${\scriptstyle (\pm0.01)}$ \\
        HGN & \textbf{0.94}${\scriptstyle (\pm0.01)}$ & 0.63${\scriptstyle (\pm0.02)}$ & 0.92${\scriptstyle (\pm0.02)}$ \\
        RGCN    & 0.91${\scriptstyle (\pm0.01)}$ & 0.6${\scriptstyle (\pm0.01)}$  & 0.9${\scriptstyle (\pm0.02)}$ \\
        GTN     & 0.9${\scriptstyle (\pm0.01)}$ & 0.62${\scriptstyle (\pm0.01)}$ & 0.91${\scriptstyle (\pm0.01)}$ \\
        FastGTN & 0.92${\scriptstyle (\pm0.00)}$ & 0.63${\scriptstyle (\pm0.01)}$ & \textbf{0.93}${\scriptstyle (\pm0.00)}$ \\
        MP-GNN   & \textbf{0.94}${\scriptstyle (\pm0.01)}$ & \textbf{0.64}${\scriptstyle (\pm0.01)}$ & \textbf{0.93}${\scriptstyle (\pm0.00)}$\\
        
        \hline
        GTN/   & APCPA,  & MAM,   & PAP,  \\
        FastGTN     & APAPA,  & MDM,   & PSP,  \\
            & APA   & MDMDM    &  \\ 
        MP-GNN & APCPA,  & MAM,   & PAP,  \\
            & APAPA   & MDM    & PSP \\ 
        \hline
    \end{tabular}
    \end{minipage}
\end{table}

Table \ref{tab:acmresults} (top) shows the $F_1$ scores achieves by the different methods. As expected, all approaches achieve similar results, which are consistent with the ones observed in previous work \cite{yun2021graph}. Indeed, the number of relations is very limited (three for IMDB, four for DBLP and ACM) and, most importantly, no relations are shared among different node pair types, substantially restricting the number of candidate meta-paths. Still, \method achieves slightly better results, most likely thanks to its ability to select a minimal set of meta-paths, as shown in Table \ref{tab:acmresults} (bottom).

\subsection{Q3: \method substantially outperforms competitors on real-world datasets with many relations}
\label{fb15k237results}

The last set of experiments aims to evaluate \method in a complex real-world setting characterized by a large set of relations, as typical of general-purpose knowledge graphs. We thus designed a set of node-classification tasks over \textbf{FB15K-237} \cite{toutanova-chen-2015-observed}, which is a large knowledge graph derived from Freebase. Each entity in the graph is associated with a text description, that we transformed into a bag-of-words representation of length 100 (retaining the most frequent words in the dataset). We identified as target labels all many-to-one relations that have from 2 to  20 possible destination types (to avoid having classes with too few examples). Examples include gender, event type and a number of currency-related relations. See the Appendix for the statistics of the datasets.

\begin{figure}[ht]
    \includegraphics[width=0.9\linewidth]{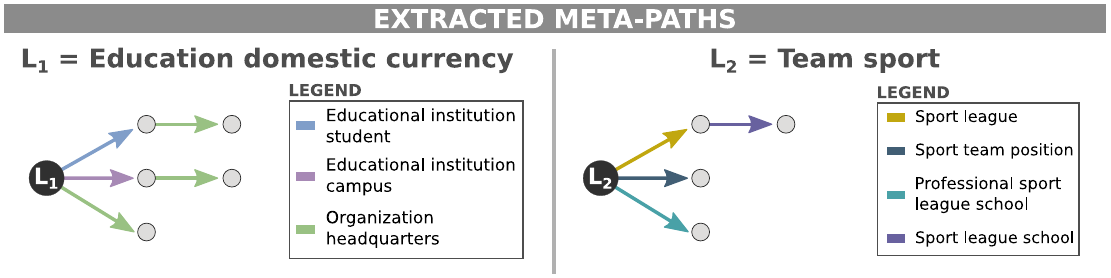}
    \caption{Examples of learned meta-paths on two node classification tasks} 

    \label{fig:new_meta_path_example}
    \end{figure}

Table \ref{tab:fb15kresults} reports $F_1$ scores for the different methods. GTN and FastGTN have serious difficulties in learning reasonable models in all cases. Indeed, the unbalancing in the class distribution, combined with the large pool of candidate relations to learn from, drives them to boil down to majority class prediction in most cases. Despite the better performance of RGCN, HGN, and R-HGNN, they still exhibit substantially lower F1-scores compared to \method. Notably \method is surpassed only by RGCN and R-HGNN in the "event" and "team sport" classification tasks, respectively.
Figure \ref{fig:new_meta_path_example} shows some examples of extracted meta-paths for two different classification tasks, namely predicting the currency of domestic tuition fees in educational institutions and predicting the sport a team is playing. In the former case, extracted meta-paths lead to the headquarters of the organization delivering the educational program, which clearly correlate with the currency being used. In the latter case, meta-paths include the league where the team is playing, which again carries information about the sport being played. Note that in both cases, node features are crucial in leveraging meta-path information, as there are not enough examples to generalize via, e.g.,  specific headquarter or sport league name. Indeed, an ablation experiment excluding node feature information (the typical setting of meta-path mining approaches~\cite{7022636,Discovering_Meta-Paths_in_Large_Heterogeneous_Information_Networks,taocohen}), shows that none of the methods manages to learn any sensible meta-path, always boiling down to learning majority class prediction rules (see Appendix~\ref{app:no_node_features}). For the same reasons, plain meta-path mining fails to extract sensible meta-paths, resulting in poor performance (see Appendix~\ref{app:mining} for the results using the popular PRA meta-path miner~\cite{taocohen}).

\begin{table}[ht]
\begin{minipage}{.24\textwidth}
    \caption{Many-relations dataset: F1-scores for the different node classification tasks on the FB15K-237 dataset. Results with standard deviations can be found in Table \ref{tab:fb15kresults_std} in Appendix. See Table~\ref{tab:DBLP} in the Appendix for the meaning of the label acronyms.
    }
    \label{tab:fb15kresults}
\end{minipage}\hfill\begin{minipage}{.75\textwidth}
    \begin{tabular}{lllllll}
        \hline
        Label & R-HGNN & HGN & RGCN & GTN & FastGTN & MP-GNN  \\ \hline
        PNC & 0.72 & 0.68 & 0.74 & 0.33  & 0.33 & \textbf{0.83} \\
        EDC & 0.6 & 0.75 & 0.71 & 0.12 & 0.12  & \textbf{0.96} \\
        EIC & 0.63 & 0.65 & 0.73 & 0.12  & 0.12 & \textbf{0.8} \\
        ELC & 0.47 & 0.74 & 0.72 & 0.12  & 0.15 & \textbf{0.78} \\
        FBC & 0.45 & 0.48 & 0.42 & 0.14  & 0.14 & \textbf{0.61} \\
        GNC & 0.8 & 0.74 & 0.82 & 0.19  & 0.19 & \textbf{0.90} \\
        OC & 0.67 & 0.73 & 0.78 & 0.14  & 0.14 & \textbf{0.93} \\
        G & 0.81 & 0.64 & 0.8 & 0.44  & 0.44 & \textbf{0.84} \\
        TS & \textbf{0.67} & 0.53 & 0.62 & 0.09  & 0.09 & 0.63 \\
        E & 0.89 & 0.8 & \textbf{0.98} & 0.07  & 0.07 & 0.96 \\
    \end{tabular}
\end{minipage}
\end{table}

Finally, to assess the computational efficiency of \method, we conducted a running time comparison, detailed in Appendix~\ref{app:exec_time}. Results show that our approach is comparable with that of the competitors on the synthetic and few relation (IMDB, DBLP, ACM) datasets. On the freebase tasks, which have a larger set of candidate relations, our approach is more expensive than (most) competitors, but these have substantially lower performance in terms of F1, with the fastest approaches (GTN and FastGTN) completely failing to learn anything sensible.

\section{Conclusion}

In this work we introduced a novel approach inspired by information theoretic principles to effectively learn meta-paths and meta-path based multi-relational GNNs in settings characterized by a large number of candidate relations combined with arbitrarily rich node features. 
Our experimental evaluation confirms the potential of the approach in recovering correct (in synthetic tasks) and informative (in real-world tasks) meta-paths despite the large number of candidate relations, a setting where existing multi-relational GNNs struggle to learn meaningful models.

Future work includes generalizing the approach to account for counts or proportions of meta-path realizations as relevant features, as well as more complex relational structures like meta-trees.

\section*{Acknowledgments}
This research was supported by TAILOR, a project funded
by the EU Horizon 2020 research and innovation program under GA No 952215. AL acknowledges the support of the MUR PNRR project FAIR - Future AI Research (PE00000013) funded by the NextGenerationEU.

\bibliographystyle{splncs04}
\bibliography{reference}

\section{Appendix}

\subsection{Weight initialization for the scoring function}
\label{sec:si:w_inizialization}
Given a relation $r$ to be evaluated for potential informativeness, node weights are initialized with values that are consistent with the assumption that $r$ determines the target values to be predicted. As discussed in Section~\ref{scorefunctiondescription}, if a node has a negative label, its neighbours according to relation $r$ cannot be part of a ground-truth meta-path, thus their weight should be close to zero. On the other hand, if all the nodes that are connected to a given node via relation $r$ have a positive label, than that given node could be part of a ground-truth meta-path. We thus initialized node weights are follows:
\begin{equation}
    w[i] = \min_{j \,:\, i \in {\cal N}_j^r} L[j] + \epsilon
\end{equation}
where $j \,:\, i \in {\cal N}_j^r$ states that node $j$ is connected to $i$ via relation $r$, $L[j]$ is the label of node $j$ and $\epsilon \in [-0.3,0.3]$ is a random value that helps breaking up symmetries during the weight learning process.

\subsection{Re-label nodes inside bags} 
\label{sec:si:relabeling}
As discussed in Section~\ref{scorefunctiondescription}, from the second iteration onwards targets are associated to bags rather than individual nodes. Adapting the procedure in Eq~\ref{eq:pos_bags} replacing node $i$ with a bag of nodes ends up diluting information too much, so that the informativeness of relations becomes difficult to predict.
On the other hand, directly using the weights learned using~\ref{eq:sf_loss} typically leads to an underestimation of the set of positive nodes, as it is sufficient that one node in the set of neighbours of a positive node $i$ has a large weight for Equation~\ref{eq:sf_first_iter} to compute the correct label for $i$. Empirically, we found that running the weight learning procedure $M$ times, with different random initializations of weights as discussed in Section~\ref{sec:si:w_inizialization}, and retaining for each node its maximum weight across runs, strikes a better balance between precision and coverage in positive node prediction, and thus a better estimate of the informativeness of the corresponding relation. We set $M=10$ in all our experiments.

\begin{figure}[ht]
	\centering
	\includegraphics[width=0.9\linewidth]{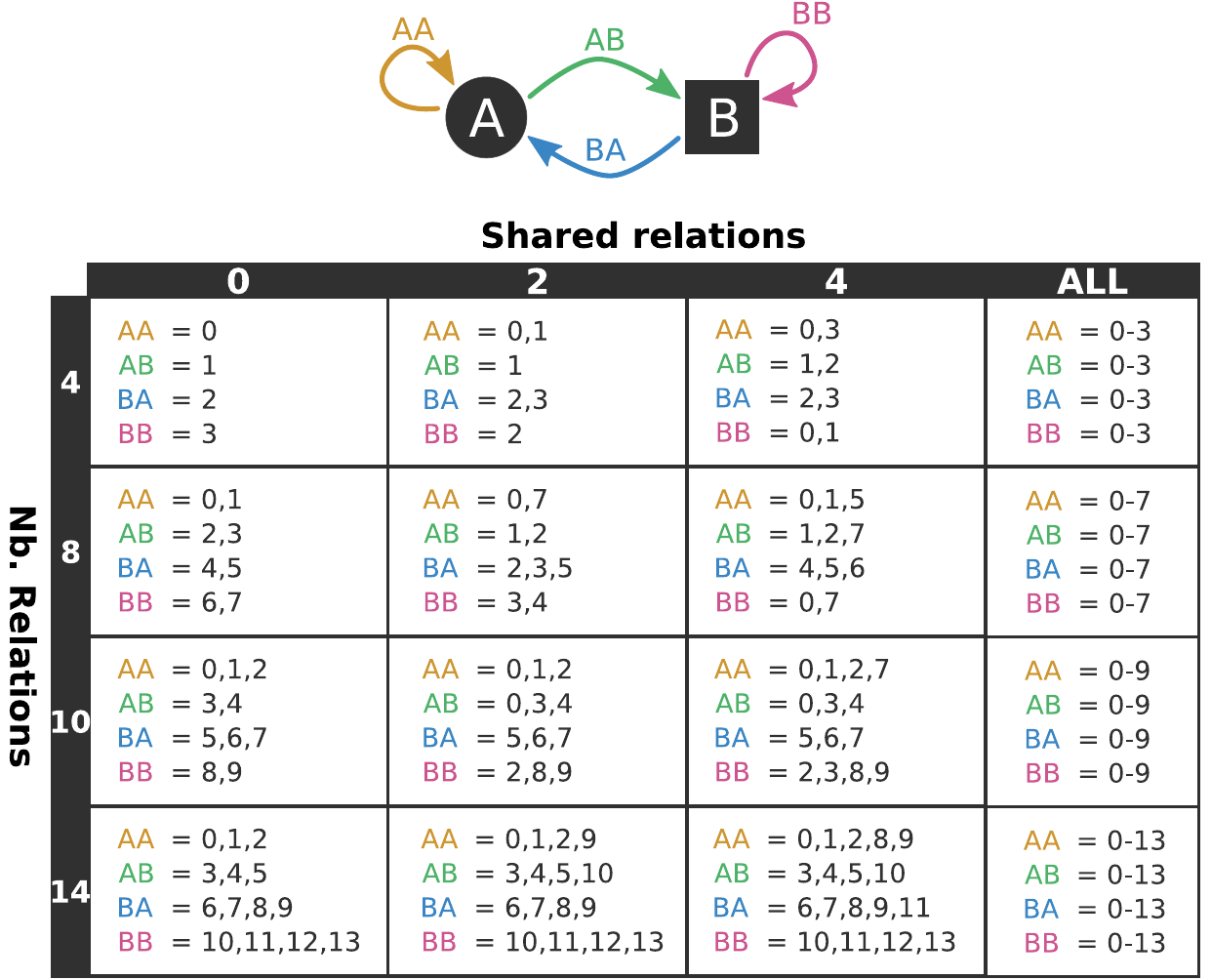}
	\caption{{\bf Synthetic experimental setting.} Each cell is a different experimental setting. $A$ and $B$ are different node types, while $AB=2,3$ indicates a setting in which $A$ nodes can be connected to $B$ nodes via relation $2$ or $3$. Moving from top to bottom we have settings with a large set of relations, while moving from left to right we have settings where relations are progressively less "pure", i.e., the same relation can connect different node-type pairs (e.g., in the second column of the first row there are 2 shared relations: relation 1 that can connect $A$ to $A$ but also $A$ to $B$, and relation 2 that can connect $B$ to $A$ but also $B$ to $B$). The last column indicates settings in which any relation is valid between any node-pair type.}
	\label{fig:experimental_setting}
\end{figure}


\begin{table}[ht]
\centering
\caption{{\bf Real-world datasets statistics.} PNC: Person Net Currency (currency associated with an individual's compensation), EDC: Education Domestic Currency (currency associated with domestic tuition fees in educational institutions), EIC: Education International Currency (currency associated with international tuition fees in educational institutions), ELC: Education Local Currency (currency associated with local tuition fees in educational institutions), FBC: Film Budget Currency (currency associated with film budget), GNC: Gdp Nominal Currency (currency associated with GDP nominal value), OC: Organization Currency (currency associated with organization), G: Gender (gender of a person), TS: Team Sport (sport of a specific team), E: Event (links recurring events to their repetitions).} 
\label{tab:DBLP}
\begin{tabular}{lllllllll}
\hline
Dataset & \# Nodes & \# Relations & \# Edges & \# Features & \# Classes & \# Target nodes  \\ \hline
DBLP    & 18405    &    4   & 67496    & 334  & 4 & 4057  \\
ACM     & 8994     &    4   & 25922    & 1902 & 3 & 3025\\
IMDB    & 12772    &    3   & 37288    & 1256 & 3 & 2939\\ \hline
FB15K-237 (PNC) & 14951 & 236 & 306711 & 100 & 3 & 583\\
FB15K-237 (EDC) & 14951 & 236 & 308201 & 100 & 6 & 481\\
FB15K-237 (EIC) & 14951 & 236 & 305559 & 100 & 6 & 119 \\
FB15K-237 (ELC) & 14951 & 236 & 307017 & 100 & 6 & 361\\
FB15K-237 (FBC) & 14951 & 236 & 315131 & 100 & 7 & 1471\\
FB15K-237 (GNC) & 14951 & 236 & 305827 & 100 & 3 & 194\\
FB15K-237 (OC) & 14951 & 236 & 307611 & 100 & 6 & 460\\
FB15K-237 (G) & 14951 & 236 & 305557 & 100 & 2 & 4530\\
FB15K-237 (TS) & 14951 & 236 & 453554 & 100 & 9  & 523\\
FB15K-237 (E) & 14951 & 236 & 310742 & 100 & 9 & 134\\
\end{tabular}
\end{table}

\clearpage
\section{Learning meta-path without node features}
\label{app:no_node_features}
In this section, we extract meta-paths using the scoring function, excluding the employment of node features. Following this, we apply the identified meta-paths to train MPGNN. A brief analysis of Table \ref{app:tab:no_node_features} clearly illustrates the crucial impact of node features in extracting valuable meta-paths. The only exception is observed in the case of synthetic dataset 1, which is unsurprising given its inherent simplicity.

\begin{table}[ht]
    \centering
    \caption{MPGNN with and without considering node features in the scoring function.}
    \label{app:tab:no_node_features}
    \begin{tabular}{ccc}
        Dataset & With NF & Without NF\\
        \midrule
        Synthetic 1 & \textbf{1}${\scriptstyle (\pm0.00)}$    & \textbf{1}${\scriptstyle (\pm0.00)}$ \\
        Synthetic 2 & \textbf{1}${\scriptstyle (\pm0.00)}$    & 0.79${\scriptstyle (\pm0.05)}$ \\
        Synthetic 3 & \textbf{1}${\scriptstyle (\pm0.00)}$    & 0.48${\scriptstyle (\pm0.03)}$ \\
        Synthetic 4 & \textbf{0.97}${\scriptstyle (\pm0.02)}$ & 0.49${\scriptstyle (\pm0.05)}$ \\
        IMDB        & \textbf{0.64}${\scriptstyle (\pm0.01)}$ & 0.55${\scriptstyle (\pm0.03)}$ \\
        DBLP        & \textbf{0.94}${\scriptstyle (\pm0.01)}$ & 0.79${\scriptstyle (\pm0.02)}$ \\
        ACM         & \textbf{0.93}${\scriptstyle (\pm0.00)}$ & 0.83${\scriptstyle (\pm0.03)}$ \\
        PNC (FB15K) & \textbf{0.83}${\scriptstyle (\pm0.01)}$ & 0.49${\scriptstyle (\pm0.03)}$ \\
        EDC (FB15K) & \textbf{0.96}${\scriptstyle (\pm0.01)}$ & 0.55${\scriptstyle (\pm0.04)}$ \\
        TS (FB15K)  & \textbf{0.63}${\scriptstyle (\pm0.03)}$ & 0.52${\scriptstyle (\pm0.01)}$ \\
        
        \bottomrule
    \end{tabular}
\end{table}

\section{Comparison with meta-path miner}
\label{app:mining}
The scoring function we propose incrementally builds relevant meta-paths. In this section, we replace the meta-paths identified by the scoring function with those obtained through conventional mining methods. In particular, we rank meta-paths using the well-known path-ranking-algorithm (PRA)\cite{taocohen}. Subsequently, we choose the top three most relevant meta-paths with lengths of 2, 3, and 4. These selected meta-paths are then used to train the MP-GNN.
In Table \ref{app:tab:mining}, you can observe the macro-F1 scores for three distinct Freebase datasets: PNC, EDC, and TS. It's important to note that we did not conduct this experiment with IMBD, DBLP, and ACM due to their limited number of relations (3 and 4). Table \ref{app:tab:mining} clearly indicates that the meta-paths identified by the scoring function exhibit significantly higher predictive capabilities compared to those obtained through PRA.

\begin{table}[ht]
    \centering
    \caption{Macro-F1 scores for three distinct Freebase datasets}
    \label{app:tab:mining}
    \begin{tabular}{lcccc}
    \toprule
    & & PNC & EDC & TS  \\ 
    \midrule
    \multirow{3}{*}{PRA + MP-GNN}  
                                 & meta-path length = 2 & 0.49 & 0.14 & 0.20 \\
                                 & meta-path length = 3 & 0.49 & 0.14 & 0.11 \\
                                 & meta-path length = 4 & 0.49 & 0.14 & 0.11 \\
    Scoring Function + MP-GNN &  & \textbf{0.83} & \textbf{0.96} & \textbf{0.63}\\

        \bottomrule
    \end{tabular}
    \end{table}

\section{Execution Time}
\label{app:exec_time}
In Table \ref{app:tab:exec_times}, we provide information on the training time for each model on individual datasets, along with the corresponding F1 scores enclosed in parentheses. Additionally, we present the average execution time and F1 score across the three benchmark datasets (IMBD, DBLP, and ACM), as well as for Freebase and synthetic datasets.
All the models achieve a similar average accuracy on the three benchmark datasets, with RGCN displaying the shortest execution time and GTN the longest. Despite being the second slowest model in terms of execution time, our model achieves the highest average F1 score. It is important to emphasize that our model is designed to learn meaningful meta-paths in networks with a multitude of relation types, whereas the three benchmark datasets only have a limited number of relation types.
In the Freebase network, both FastGTN and GTN boast the shortest execution times, yet their F1 scores are significantly lower, rendering them ineligible for further consideration. 
Despite Simple-HGN being the quickest model, its F1 score falls significantly short, averaging 15 points lower than MP-GNN. On the other hand, RGCN is the second fastest model, but once more, its F1 score lags behind our approach by an average of 9 points.
Lastly, concerning the synthetic datasets, our model ranks third in terms of execution time, with Simple-HGN excluded from the analysis due to its poor F1 scores. It's worth highlighting that the fastest model, FastGTN, and the second fastest, RGCN, both lag behind MP-GNN by 23 and 17 points, respectively.
In general, our approach is consistently neither the slowest nor the fastest; however, it consistently attains the highest average F1 score in all scenarios.

\begin{table}[ht]
    \centering
    \caption{Execution time in seconds, with the F1 score in parentheses.}
    \label{app:tab:exec_times}
\begin{tabular}{lcccccc}        
\toprule
 &R-HGNN & Simple-HGN & RGCN & GTN & FastGTN & MP-GNN \\
        \midrule
        IMDB & 680 ${\scriptstyle (0.64)}$ & 740 ${\scriptstyle (0.63)}$ & 650 ${\scriptstyle (0.60)}$ & 1500 ${\scriptstyle (0.62)}$ & 910 ${\scriptstyle (0.63)}$ & 1000 ${\scriptstyle (0.64)}$\\
        DBLP & 720 ${\scriptstyle (0.86)}$ & 870 ${\scriptstyle (0.94)}$ & 780 ${\scriptstyle (0.91)}$ & 1630 ${\scriptstyle (0.90)}$ & 990 ${\scriptstyle (0.92)}$ & 1180 ${\scriptstyle (0.94)}$\\
        ACM & 940 ${\scriptstyle (0.90)}$ & 750 ${\scriptstyle (0.92)}$ & 870 ${\scriptstyle (0.90)}$ & 1420 ${\scriptstyle (0.91)}$ & 870 ${\scriptstyle (0.93)}$ & 960 ${\scriptstyle (0.93)}$\\
        \hline
\textbf{Mean} & 780 ${\scriptstyle (0.80)}$ & 786 ${\scriptstyle (0.83)}$ & 767 ${\scriptstyle (0.80)}$ & 1517 ${\scriptstyle (0.81)}$ & 923 ${\scriptstyle (0.83)}$ & 1046 ${\scriptstyle (0.84)}$\\
        \hline\\ \\
PNC & 7230 ${\scriptstyle (0.72)}$ & 560 ${\scriptstyle (0.68)}$ & 1830 ${\scriptstyle (0.74)}$ & 180 ${\scriptstyle (0.33)}$ & 150 ${\scriptstyle (0.33)}$ & 2870 ${\scriptstyle (0.83)}$\\
EDC & 7356 ${\scriptstyle (0.60)}$ & 670 ${\scriptstyle (0.75)}$ & 1540 ${\scriptstyle (0.71)}$ & 190 ${\scriptstyle (0.12)}$ & 130 ${\scriptstyle (0.12)}$ & 3220 ${\scriptstyle (0.96)}$\\
EIC & 7020 ${\scriptstyle (0.63)}$ & 460 ${\scriptstyle (0.65)}$ & 2040 ${\scriptstyle (0.73)}$ & 180 ${\scriptstyle (0.12)}$ & 110 ${\scriptstyle (0.12)}$ & 2480 ${\scriptstyle (0.80)}$\\
ELC & 820 ${\scriptstyle (0.47)}$ & 760 ${\scriptstyle (0.74)}$ & 1380 ${\scriptstyle (0.72)}$ & 180 ${\scriptstyle (0.15)}$ & 130 ${\scriptstyle (0.15)}$ & 3010 ${\scriptstyle (0.78)}$\\
FBC & 5900 ${\scriptstyle (0.45)}$ & 410 ${\scriptstyle (0.48)}$ & 1190 ${\scriptstyle (0.42)}$ & 190 ${\scriptstyle (0.14)}$ & 100 ${\scriptstyle (0.14)}$ & 2880 ${\scriptstyle (0.60)}$\\
GNC & 8230 ${\scriptstyle (0.80)}$ & 670 ${\scriptstyle (0.74)}$ & 1680 ${\scriptstyle (0.82)}$ & 175 ${\scriptstyle (0.19)}$ & 100 ${\scriptstyle (0.19)}$ & 3220 ${\scriptstyle (0.90)}$\\
OC & 3790 ${\scriptstyle (0.67)}$ & 670 ${\scriptstyle (0.73)}$ & 1970 ${\scriptstyle (0.78)}$ & 185 ${\scriptstyle (0.14)}$ & 120 ${\scriptstyle (0.14)}$ & 2980 ${\scriptstyle (0.93)}$\\
G & 5980 ${\scriptstyle (0.81)}$ & 450 ${\scriptstyle (0.64)}$ & 2010 ${\scriptstyle (0.80)}$ & 140 ${\scriptstyle (0.44)}$ & 120 ${\scriptstyle (0.44)}$ & 2990 ${\scriptstyle (0.84)}$\\
TS & 5000 ${\scriptstyle (0.67)}$ & 410 ${\scriptstyle (0.53)}$ & 1995 ${\scriptstyle (0.62)}$ & 200 ${\scriptstyle (0.09)}$ & 120 ${\scriptstyle (0.09)}$ & 3155 ${\scriptstyle (0.63)}$\\
E & 6790 ${\scriptstyle (0.89)}$ & 690 ${\scriptstyle (0.80)}$ & 2005 ${\scriptstyle (0.98)}$ & 170 ${\scriptstyle (0.07)}$ & 140 ${\scriptstyle (0.07)}$ & 3040 ${\scriptstyle (0.96)}$\\
        \hline
\textbf{Mean} & 5812 ${\scriptstyle (0.67)}$ & 575 ${\scriptstyle (0.67)}$ & 1764 ${\scriptstyle (0.73)}$ & 179 ${\scriptstyle (0.18)}$ & 122 ${\scriptstyle (0.18)}$ & 2984 ${\scriptstyle (0.82)}$\\
        \hline\\ \\
        
Synt 1 & 320 ${\scriptstyle (1.00)}$ & 50 ${\scriptstyle (034)}$ & 200 ${\scriptstyle (1.00)}$ & 230 ${\scriptstyle (0.93)}$ & 210 ${\scriptstyle (0.94)}$ & 245 ${\scriptstyle (1.00)}$\\
Synt 2 & 310 ${\scriptstyle (1.00)}$ & 66 ${\scriptstyle (0.48)}$ & 240 ${\scriptstyle (0.91)}$ & 210 ${\scriptstyle (0.84)}$ & 180 ${\scriptstyle (0.85)}$ & 300 ${\scriptstyle (1.00)}$\\
Synt 3 & 350 ${\scriptstyle (1.00)}$ & 68 ${\scriptstyle (0.38)}$ & 300 ${\scriptstyle (0.95)}$ & 310 ${\scriptstyle (0.84)}$ & 280 ${\scriptstyle (0.81)}$ & 345 ${\scriptstyle (0.99)}$\\
Synt 4 & 410 ${\scriptstyle (1.00)}$ & 78 ${\scriptstyle (0.50)}$ & 390 ${\scriptstyle (0.84)}$ & 480 ${\scriptstyle (0.47)}$ & 320 ${\scriptstyle (0.52)}$ & 430 ${\scriptstyle (1.00)}$\\
Synt 5 & 450 ${\scriptstyle (0.98)}$ & 67 ${\scriptstyle (0.34)}$ & 400 ${\scriptstyle (0.85)}$ & 400 ${\scriptstyle (0.91)}$ & 360 ${\scriptstyle (0.94)}$ & 380 ${\scriptstyle (1.00)}$\\
Synt 6 & 430 ${\scriptstyle (0.95)}$ & 120 ${\scriptstyle (0.49)}$ & 390 ${\scriptstyle (1.00)}$ & 460 ${\scriptstyle (0.82)}$ & 400 ${\scriptstyle (0.84)}$ & 450 ${\scriptstyle (1.00)}$\\
Synt 7 & 450 ${\scriptstyle (0.96)}$ & 94 ${\scriptstyle (0.52)}$ & 450 ${\scriptstyle (0.68)}$ & 500 ${\scriptstyle (0.8)}$ & 490 ${\scriptstyle (0.8)}$ & 480 ${\scriptstyle (1.00)}$\\
Synt 8 & 520 ${\scriptstyle (0.89)}$ & 128 ${\scriptstyle (0.50)}$ & 460 ${\scriptstyle (0.71)}$ & 720 ${\scriptstyle (0.47)}$ & 450 ${\scriptstyle (0.48)}$ & 440 ${\scriptstyle (1.00)}$\\
Synt 9 & 560 ${\scriptstyle (0.93)}$ & 135 ${\scriptstyle (0.38)}$ & 425 ${\scriptstyle (0.90)}$ & 530 ${\scriptstyle (0.90)}$ & 430 ${\scriptstyle (0.92)}$ & 510 ${\scriptstyle (1.00)}$\\
Synt 10 & 590 ${\scriptstyle (0.89)}$ & 90 ${\scriptstyle (0.45)}$ & 580 ${\scriptstyle (0.85)}$ & 600 ${\scriptstyle (0.87)}$ & 520 ${\scriptstyle (0.86)}$ & 540 ${\scriptstyle (1.00)}$\\
Synt 11 & 630 ${\scriptstyle (0.85)}$ & 200 ${\scriptstyle (0.64)}$ & 495 ${\scriptstyle (0.77)}$ & 640 ${\scriptstyle (0.8)}$ & 510 ${\scriptstyle (0.79)}$ & 500 ${\scriptstyle (0.94)}$\\
Synt 12 & 630 ${\scriptstyle (0.90)}$ & 130 ${\scriptstyle (0.36)}$ & 500 ${\scriptstyle (0.62)}$ & 605 ${\scriptstyle (0.48)}$ & 525 ${\scriptstyle (0.48)}$ & 560 ${\scriptstyle (1.00)}$\\
Synt 13 & 610 ${\scriptstyle (0.82)}$ & 175 ${\scriptstyle (0.49)}$ & 565 ${\scriptstyle (0.83)}$ & 670 ${\scriptstyle (0.88)}$ & 570 ${\scriptstyle (0.85)}$ & 580 ${\scriptstyle (1.00)}$\\
Synt 14 & 650 ${\scriptstyle (0.88)}$ & 200 ${\scriptstyle (0.32)}$ & 595 ${\scriptstyle (0.81)}$ & 650 ${\scriptstyle (0.8)}$ & 600 ${\scriptstyle (0.84)}$ & 560 ${\scriptstyle (1.00)}$\\
Synt 15 & 600 ${\scriptstyle (0.87)}$ & 185 ${\scriptstyle (0.42)}$ & 650 ${\scriptstyle (0.84)}$ & 710 ${\scriptstyle (0.77)}$ & 680 ${\scriptstyle (0.8)}$ & 590 ${\scriptstyle (1.00)}$\\
Synt 16 & 660 ${\scriptstyle (0.81)}$ & 200 ${\scriptstyle (0.45)}$ & 700 ${\scriptstyle (0.6)}$ & 720 ${\scriptstyle (0.48)}$ & 705 ${\scriptstyle (0.48)}$ & 605 ${\scriptstyle (0.97)}$\\
        \hline

\textbf{Mean} & 523 ${\scriptstyle (0.92)}$ & 129 ${\scriptstyle (0.44)}$ & 476 ${\scriptstyle (0.82)}$ & 547 ${\scriptstyle (0.75)}$ & 468 ${\scriptstyle (0.76)}$ & 484 ${\scriptstyle (0.99)}$\\
        
        \bottomrule
\end{tabular}
\end{table}

\clearpage
\section{FB15K-237 Results with standard deviations}
\begin{table}[ht]
    \caption{Many-relations dataset: F1-scores for the different node classification tasks on the FB15K-237 dataset. Mean and standard deviation computed over five runs with different random seeds. Best results are highlighted in bold. See Table~\ref{tab:DBLP} in the Appendix for the meaning of the label acronyms.}
    \label{tab:fb15kresults_std}
    \centering
    \begin{tabular}{lllllll}
            \hline
            Label & R-HGNN & HGN & RGCN & GTN & FastGTN & MP-GNN  \\ \hline
            PNC & 0.72${\scriptstyle (\pm0.04)}$ & 0.68${\scriptstyle (\pm0.05)}$ & 0.74${\scriptstyle (\pm0.03)}$ & 0.33${\scriptstyle (\pm0.00)}$  & 0.33${\scriptstyle (\pm0.00)}$ & \textbf{0.83}${\scriptstyle (\pm0.01)}$ \\
            EDC & 0.6${\scriptstyle (\pm0.05)}$ & 0.75${\scriptstyle (\pm0.02)}$ & 0.71${\scriptstyle (\pm0.02)}$ & 0.12${\scriptstyle (\pm0.00)}$ & 0.12${\scriptstyle (\pm0.00)}$  & \textbf{0.96} ${\scriptstyle (\pm0.01)}$\\
            EIC & 0.63${\scriptstyle (\pm0.06)}$ & 0.65${\scriptstyle (\pm0.04)}$ & 0.73${\scriptstyle (\pm0.01)}$ & 0.12${\scriptstyle (\pm0.00)}$  & 0.12${\scriptstyle (\pm0.00)}$ & \textbf{0.8}${\scriptstyle (\pm0.03)}$ \\
            ELC & 0.47${\scriptstyle (\pm0.07)}$ & 0.74${\scriptstyle (\pm0.01)}$ & 0.72${\scriptstyle (\pm0.02)}$ & 0.12${\scriptstyle (\pm0.00)}$  & 0.15${\scriptstyle (\pm0.00)}$ & \textbf{0.78}${\scriptstyle (\pm0.02)}$ \\
            FBC & 0.45${\scriptstyle (\pm0.04)}$ & 0.48${\scriptstyle (\pm0.02)}$ & 0.42${\scriptstyle (\pm0.00)}$ & 0.14${\scriptstyle (\pm0.00)}$  & 0.14${\scriptstyle (\pm0.00)}$ & \textbf{0.61}${\scriptstyle (\pm0.01)}$ \\
            GNC & 0.8${\scriptstyle (\pm0.05)}$ & 0.74${\scriptstyle (\pm0.02)}$ & 0.82${\scriptstyle (\pm0.03)}$ & 0.19${\scriptstyle (\pm0.00)}$  & 0.19${\scriptstyle (\pm0.00)}$ &  \textbf{0.9}${\scriptstyle (\pm0.00)}$\\
            OC & 0.67${\scriptstyle (\pm0.02)}$ & 0.73${\scriptstyle (\pm0.04)}$ & 0.78${\scriptstyle (\pm0.01)}$ & 0.14${\scriptstyle (\pm0.00)}$  & 0.14${\scriptstyle (\pm0.00)}$ &  \textbf{0.93}${\scriptstyle (\pm0.01)}$\\
            G & 0.81${\scriptstyle (\pm0.01)}$ & 0.64 ${\scriptstyle (\pm0.01)}$ & 0.8${\scriptstyle (\pm0.01)}$ & 0.44${\scriptstyle (\pm0.00)}$  & 0.44${\scriptstyle (\pm0.00)}$ &  \textbf{0.84}${\scriptstyle (\pm0.04)}$\\
            TS & \textbf{0.67}${\scriptstyle (\pm0.04)}$ & 0.53${\scriptstyle (\pm0.04)}$ & 0.62${\scriptstyle (\pm0.01)}$ & 0.09${\scriptstyle (\pm0.00)}$  & 0.09${\scriptstyle (\pm0.00)}$ &  0.63${\scriptstyle (\pm0.03)}$ \\
            E & 0.89${\scriptstyle (\pm0.02)}$ & 0.8${\scriptstyle (\pm0.03)}$ & \textbf{0.98}${\scriptstyle (\pm0.00)}$ & 0.07${\scriptstyle (\pm0.00)}$  & 0.07${\scriptstyle (\pm0.00)}$ &  0.96${\scriptstyle (\pm0.00)}$\\
    \end{tabular}
    
\end{table}

\end{document}